\documentclass[11pt]{article}
\usepackage{eamt26}
\usepackage{times}
\usepackage{url}
\usepackage{latexsym}
\usepackage[small,bf]{caption} 
\title{A Technical Curriculum on Language-Oriented Artificial Intelligence in Translation and Specialised Communication}

\author{Ralph Krüger\\
  Institute of Translation and Multilingual Communication\\
  TH Köln – University of Applied Sciences Cologne, Germany\\
    {\tt ralph.krueger@th-koeln.de}}

\date{}

\begin{document}
\maketitle
\begin{abstract}
  This paper presents a technical curriculum on language-oriented artificial intelligence (AI) in the language and translation (L\&T) industry. The curriculum aims to foster domain-specific technical AI literacy among stakeholders in the fields of translation and specialised communication by exposing them to the conceptual and technical/algorithmic foundations of modern language-oriented AI in an accessible way. The core curriculum focuses on 1) vector embeddings, 2) the technical foundations of neural networks, 3) tokenization and 4) transformer neural networks. It is intended to help users develop computational thinking as well as algorithmic awareness and algorithmic agency, ultimately contributing to their digital resilience in AI-driven work environments. The didactic suitability of the curriculum was tested in an AI-focused MA course at the Institute of Translation and Multilingual Communication at TH Köln. Results suggest the didactic effectiveness of the curriculum, but participant feedback indicates that it should be embedded into higher-level didactic scaffolding – e.g., in the form of lecturer support – in order to enable optimal learning conditions. 
\end{abstract}

\section{Introduction}

The recent emergence of general-purpose AI (GPAI) technologies in the form of large language models (LLMs) has widened the scope of automation in the L\&T industry to a considerable extent. For example, a single LLM can perform translation-technological tasks that were originally distributed over different expert technologies (e.g., intra- and interlingual machine translation, terminology extraction or quality estimation). Furthermore, LLMs can augment traditional translation technologies with new features that were previously beyond the scope of these technologies (e.g., augmenting traditional automatic translation quality control with meaning-based source- and target-segment comparisons) and they can (partially) automate tasks which largely resisted being automated to date (e.g., autonomous text production at high proficiency levels or intersemiotic machine translation/interpreting).\\ The current LLM-driven automation cycle in the L\&T industry is shaping up to be both more extensive and more invasive than previous automation cycles, as evidenced, for example, by the \textit{LangOps Manifesto}'s \textit{AI-first} principle.\footnote{``AI systems are already good enough for many use cases and their performance will continue to improve. We always start by trying AI solutions to solve language problems.`` (\url{https://langops.org/our-manifesto/})} It could be claimed that – in order to retain high levels of agency and control in AI-first environments in the spirit of the human-centered AI approach \cite{Shneiderman_2020} – L\&T industry stakeholders require domain-specific AI literacy, as conceptualised, e.g., in the AI literacy framework proposed in Krüger \shortcite{Krueger_2024} and Krüger \shortcite{Krueger_forthcoming}. In this framework, overarching AI literacy is further decomposed into the individual subdimensions of technical, performance-focused, interaction-focused, implementation-focused and ethical/societal AI literacy, which are interrelated in various ways.\\
The curriculum proposed in this paper focuses on developing technical AI literacy, which involves knowledge of the basic operating principles of modern – currently to a large extent transformer-based – AI technologies. It is certainly true that, as noted by Bowker \shortcite[73]{Bowker_2026} with regard to translation didactics, ``[i]t is not necessary for most translation educators or students to understand all the details of how an artificial neural network operates in order to use AI-based translation.`` This view is complemented by Rivas Ginel and Moorkens \shortcite[295]{Rivas_Ginel_Moorkens_2025}, who found that translators' (dis)trust in AI technologies such as LLMs is mostly calibrated by their use of these systems rather than by a deeper understanding of the systems' technical foundations.\\ However, there are also several arguments in favour of fostering technical AI literacy in L\&T industry stakeholders. Most importantly, technical AI literacy has an element of empowerment in that it helps in demystifying the black-box nature of modern machine-learning based AI technologies such as as LLMs, a point which was previously made by Doherty and Kenny \shortcite[296]{Doherty_Kenny_2014} in the context of statistical machine translation (SMT) and by Kenny \shortcite[438]{Kenny_2019} in the context of neural machine translation (NMT). AI-literate stakeholders familiar with the operating principles of these technologies may have a better-informed picture of the tasks these technologies can currently be used for and the level of quality at which these tasks can be completed (aspects related to performance- and implementation-focused AI literacy). This may also enable them to assume technology-oriented consulting roles in L\&T industry contexts; see, e.g., the ``consulting competence`` in the post-editing competence model proposed by Nitzke et al. \shortcite[248]{Nitzke_et_al_2019}. At a higher level of abstraction, technical AI literacy may also help L\&T stakeholders develop their computational thinking \cite[4]{Celik_2023} and, related to this, their algorithmic awareness \cite{Gran_et_al_2021} and algorithmic agency \cite[42]{Mills_Gutierrez_2026}, ultimately contributing to their digital resilience \cite[63, 65]{kornacki_hybrid_2024} in highly automated AI-first environments. These arguments in favour of technical AI literacy are supported, to some extent, by feedback on the didactic suitability of the technical curriculum as discussed in Section 3.3.\\
With its L\&T stakeholder focus, the curriculum stands in a tradition of initiatives such as MultiTraiNMT \cite{Kenny_2022}, DataLit\textsuperscript{MT} \cite{krueger_hackenbuchner_2024}, adaptMLLM \cite{Lankford_et_al_2023} and LT-LiDER \cite{moorkens-etal-2024-lt-lider}.

\section{The curriculum}

The technical curriculum on language-oriented AI in translation and specialised communication is provided as a GitHub repository\footnote{\url{https://github.com/ITMK/AI_Literacy/}}, which consists of a series of Jupyter notebooks provided under a CC BY-SA 4.0 License (Attribution-ShareAlike 4.0 International) and hosted in Google's Colaboratory (Colab) online environment. The core curriculum presented below is complete. Further notebooks addressing additional aspects of language-oriented AI relevant to the L\&T industry will be added to the curriculum in the future. 

\subsection{Structure of the curriculum}

The core curriculum consists of four sections, each covered by a separate Jupyter notebook. These notebooks cover 1) vector embeddings, 2) the technical
foundations of neural networks, 3) tokenization (with a focus on subword tokenization) and 4) the inner workings of transformer neural networks.
\subsubsection*{Vector embeddings – Representing linguistic data as numbers}The first notebook is concerned with vector embeddings and illustrates to users in a detailed way how linguistic data can be represented as numbers to be processed by neural networks. The notebook covers static/decontextualized and dynamic/contextualized word embeddings as well as sentence embeddings and multilingual embeddings. Users can load a pre-trained static word embedding model and train a small word embedding model themselves, explore individual word and sentence embeddings and embedding relations through visualisations and by calculating the Euclidean Distance and the Cosine Similarity between individual embeddings. The notebook also illustrates how transformer language models – taking BERT \cite{devlin-etal-2019-bert} as an example – start with static/decontextualized embeddings in their initial embedding layer and convert these into dynamic/contextualized embeddings in their output layer (this process is explored in more detail in the fourth notebook concerned with transformer language models). This notebook is intended to meet L\&T stakeholders where they are in their capacity as (aspiring) experts in natural language communication and guides them towards developing a more computational perspective on natural language that can be broadened and deepened in the further sections of the curriculum.\\
To rephrase this approach in didactic terms, the first notebook leads users ``from the zone of current development to the zone of proximal development – the space where one cannot quite master a content/task of their own, but they can with the help of an expert`` \cite[33]{Angelone_2026}. The content covered in the subsequent notebooks is then mostly situated in this zone of proximal development. The expert guidance required in this zone can be embedded into the notebooks, to some extent, through adequate didactic scaffolding as discussed in Section 2.2. If the notebooks are used as learning resources in an instructor-led course, the lecturer can provide further expert guidance (see also the discussion in Section 3.2). 

\subsubsection*{Technical foundations of neural networks – Building blocks, forward and backward pass}
The second notebook addresses the technical foundations of neural networks and introduces users to the basic building blocks of these networks (neurons, trainable parameters, activation functions), their structure (input layer, hidden layers, output layer) and the flow of information through these networks (forward and backward pass). In this notebook, which intentionally favours didactic simplicity and accessibility over functional completeness, users can create a small experimental neural network in Python by defining the individual layers and the mathematical operations performed within and between these layers. Users can then simulate a simplified forward pass through this neural network, which illustrates how such a network may perform a translation of a short example sentence. The notebook also illustrates through a simplified backward pass how, in neural network training, the network’s parameters (weights and biases) would be updated via backpropagation in order to reduce network loss. Building upon the previous notebook, users learn how word vectors are combined into a matrix and how this matrix is processed by the individual layers of the neural network. The notebook is intended to further develop users’ computational thinking and algorithmic awareness by illustrating how natural language is actually processed and how natural language semantics is represented within a neural network. At the same time, it introduces users to foundational machine learning concepts and operations such as matrices, tensors, dot-product calculation, non-linear and softmax activation functions etc., which they will encounter again in the context of transformer neural networks.

\subsubsection*{Tokenization – Reducing the vocabulary size of neural networks}
The third notebook is concerned with tokenization, which is employed in language models in order to reduce the size of the vocabulary to be learned. The notebook first guides users through word- and character-based tokenization and discusses the advantages and disadvantages of each method. From there, the notebook guides users through the process of subword-based tokenization, which combines the advantages of word- and character-based tokenization and is the tokenization method actually used in state-of-the-art language models. In this context, users are introduced to three major subword tokenization algorithms, i.e., Byte-Pair Encoding (BPE) \cite{Gage_1994}, WordPiece \cite{Schuster_Nakajima} and Unigram \cite{kudo-2018-subword}. Users can then load a BPE, a WordPiece and a Unigram tokenizer, have them each tokenize an example sentence and explore how the three algorithms differ in how they tokenize words and which special characters they employ to designate word boundaries or intra-word splits. Then, users are introduced to the concept of token IDs, which are unique identifiers associated with each token in a language model. Armed with this knowledge, users can access the vocabulary of the language model GPT-2 \cite{Radford2019LanguageMA} and explore which tokens are stored in this vocabulary. Finally, the notebook illustrates the steps involved in going from subword tokenization to vector embeddings to be processed by a language model.\\
Tokenization is covered after word embeddings and the technical foundations of neural networks since particularly subword tokenization introduces an additional level of complexity which may be confusing to users at first, but which should be less daunting once they are familiar with word embeddings and how these are processed by neural networks and once their computational thinking has evolved accordingly. Also, having encountered the encoder-only transformer model BERT in the first notebook and the decoder-only transformer model GPT-2\footnote{The curriculum relies on older language models such as BERT and GPT-2 because they are functionally less complex than current state-of-the-art models and, more importantly, because they are considerably smaller and can thus be loaded into a Google Colab environment and used in this environment at reasonable speed.} in the current notebook, users are well-equipped to proceed to the next notebook, specifically concerned with transformer language models.

\subsubsection*{Transformer neural networks}
The fourth notebook builds upon the foundations laid in the previous three notebooks and takes a deep dive into the inner workings of the transformer neural network architecture. Initial focus is placed on encoder-decoder vs. encoder-only vs. decoder-only models and on their respective areas of application and training regimes. In this context, the notebook illustrates how the Hugging Face transformers pipeline\footnote{\url{https://huggingface.co/docs/transformers/main_classes/pipelines}} defaults to encoder-decoder models for sequence2sequence tasks such as text summarization or translation, to encoder-only models for natural language understanding tasks such as named entity recognition or question answering and to decoder-only models for natural language generation tasks such as text generation/completion. Then, the notebook explores in detail the individual components of the encoder and the decoder side of an encoder-decoder transformer language model. Here, the focus is placed on the original transformer architecture proposed by Vaswani et al. \shortcite{Vaswani_2017} in the context of NMT research because, a) current LLMs are still strikingly similar to this original transformer architecture \cite{Raschka_2025a} and b) early transformer models are a bit simpler to understand than current LLMs “but still complex enough to give you a solid grasp of how modern transformer models work” \cite{Raschka_2025b}. When more profound differences exist between the original transformer architecture and current LLMs (e.g., in terms of how position embeddings are created, which activation functions are used or how layer normalization is performed), the notebook provides an `architecture update alert`, where it briefly discusses the current techniques and points users to relevant literature. On the encoder side, the notebook first illustrates how inputs to the transformer are embedded (here, users can draw on their previous knowledge acquired in the notebook on vector embeddings) and how these embeddings are enriched with positional encoding vectors. Then, the notebook explores in detail the transformer self-attention process (splitting the original input embedding representation into query, key and value representations, dot-product calculation, scaling, padding/masking, softmax normalization etc.), which creates a dynamic/contextualized representation of the initially static/decontextualized input embedding representation. Armed with this conceptual and technical/algorithmic knowledge of transformer self-attention, users can then visualize and explore the self-attention process on their own by using the BertViz package \cite{Vig_2019}. The discussion of the decoder side is more concise because the decoder is conceptually similar to the encoder, meaning that users will already be familiar with many of the components discussed here. Particular focus is placed on masked multi-head attention and on output generation in the final linear and softmax layers of the decoder. Here, users can have GPT-2 complete a text and visualize the top three token probabilities at each generation step. They can also access the user-friendly Hugging Face Decoding Visualizer\footnote{\url{https://huggingface.co/spaces/agents-course/decoding_visualizer}} to further explore greedy decoding and the Hugging Face Beam Search Visualizer\footnote{\url{https://huggingface.co/spaces/m-ric/beam_search_visualizer}} to explore the more complex beam search decoding algorithm.\\
This core curriculum is currently being complemented by further notebooks covering training and adaptation strategies for language models as well as more recent developments in language-oriented AI, such as multimodal LLMs\footnote{Including models capable of processing spoken language, which is intended to extend the applicability of the curriculum to the field of interpreting as well.}, large reasoning and action models, small language models or knowledge-enhanced LLMs (e.g., via RAG or Knowledge Graphs). The notebooks of the core curriculum are also updated on an ongoing basis according to user feedback.

\subsection{Didactic makeup of the Jupyter notebooks}
The curriculum explores the didactic potential of Jupyter notebooks in introducing audiences with little to no programming knowledge and correspondingly low levels of computational thinking and algorithmic awareness to complex concepts and algorithmic processes from fields such as natural language processing, as investigated in a translation context in Krüger \shortcite{Krueger_2022}. Jupyter notebooks support \textit{literate computing} \cite{Millman_Perez_2018} through a combination of (multimodal) documentation sections providing conceptual information interleaved with executable code cells, which illustrate the algorithmic implementation of these concepts. This combination of documentation and executable code allows authors to use these notebooks to ``tell an interactive, computational story`` \cite[7]{Barba_et_al_2019} to their readers, with a didactic scaffolding tailored to specific target audiences.The target audience \textit{L\&T industry stakeholders} as referred to in this paper includes, e.g., students of MA programmes in translation, specialised communication (as central L\&T industry fields) and related areas as well as professionals working in the L\&T industry. It is assumed that these stakeholders have a professional interest in language-oriented AI but have not been exposed in detail to the technical foundations of these technologies. It is further assumed that these stakeholders have had moderate to no exposure to computer programming before, making them novices or advanced beginners in terms of computational thinking.\\ 
Figure 1 depicts a typical example of the didactic makeup of the Jupyter notebooks within the technical curriculum. The left screenshot is concerned with a forward pass of a simple example sentence through an experimental translation neural network and illustrates how the `semantics` of a particular input word is processed and represented by a hidden layer in the network (excluding the activation function at this point). The right screenshot is concerned with self-attention in a transformer language model. The code cells implement BertViz to visualise the BERT model's self-attention process performed on a predefined example sentence (which users are encouraged to change in order to experiment with their own sentences). The documentation sections explain how to interpret the BertViz visualisation. Short exercises then invite users to engage more deeply with the topic, in the current example also highlighting points of contact with topics covered at a previous point in the curriculum.

\begin{figure*}
 \centering
 \includegraphics[width=\textwidth]{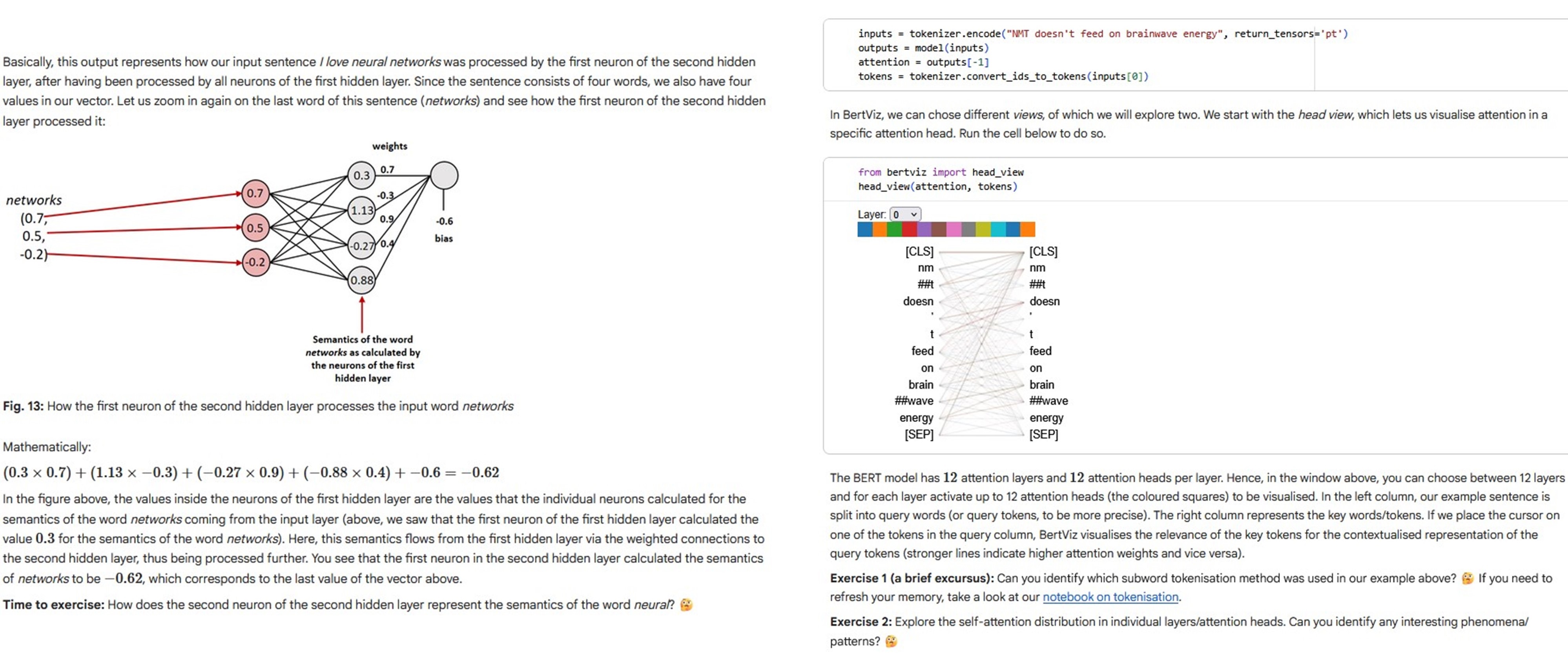}
 \caption{Example of the didactic makeup of the Jupyter notebooks – left: processing of embeddings in a hidden network layer (excluding activation); right: visualising and exploring self-attention in a transformer language model}
\end{figure*}

\section{Measuring the didactic suitability of the curriculum}
The core curriculum presented above was tested in the MA course ``Foundations of Artificial Intelligence for Specialised Communication`` in the MA in Multilingual Specialised Communication and Specialised Translation programme (code: \textit{MAFKÜ}) and the MA in Terminology and Language Technology programme (code: \textit{MATS}) at TH Köln's Institute of Translation and Multilingual Communication. The course was intended for first-year students of the two MA programmes and was also attended by three external guest students who are experienced professional translators working in public-sector language services (code: \textit{professionals}). Participants completed a pre-test questionnaire (October 2025) and a post-test questionnaire (January 2026). The study was conducted in accordance with TH Köln's Regulations for Safeguarding Good Scientific Practice\footnote{\url{https://www.th-koeln.de/mam/bilder/forschung/2023-05-31_gwp_en.pdf}} and Code of Conduct for Research and Transfer\footnote{\url{https://www.th-koeln.de/mam/bilder/forschung/verhaltenskodex_maerz_2021_en-us.pdf}}. The questionnaires were derived from the draft version of the TransAI Literacy Scale (TrAILS) proposed in Krüger \shortcite{Krueger_forthcoming} and were administered in German.\footnote{Draft version of TrAILS: \url{https://forms.gle/QdhPiP9D3F6rUca4A}\\Pre-test questionnaire: \url{https://forms.gle/VNwL8LpNdsKh6HLd8}\\Post-test questionnaire: \url{https://forms.gle/VBzvcFub3EtSLSAt6}}\\
The pre-test questionnaire was completed by 24 participants. Of these, 12 (50\%) were MAFKÜ students, 9 (37.5\%) were MATS students and 3 (12.5\%) were professionals (item 0.3). The latter group had a professional experience of more than 20 years, while the professional experience of the MATS and MAFKÜ students varied between \textless1 year (66.7\%), 1 to \textless5 years (12.5\%), 5 to \textless10 years (4.2\%) and 10 to \textless20 years (4.2\%) (item 0.2). The post-test questionnaire was completed by 15 participants, resulting in a survey attrition rate of 37.5\%. The composition of the post-test group shifted to 6 MATS students (40\%), 5 MAFKÜ students (33.3\%), 3 in-house professionals (20\%, this group remained constant) and 1 freelance professional (6.7\%).\footnote{Since the study did not use constant participant IDs for pre- and post-test, it is unclear whether the participant who identified as a freelance professional in the post-test questionnaire identified as a MAFKÜ or a MATS student in the pre-test questionnaire (both questionnaires asked participants for their `primary role` [item 0.3]; no further external guest students joined the course between the pre-test and the post-test).}\\ This shift is relevant for interpreting the results: The MATS curriculum includes a higher share of IT-related courses than the MAFKÜ curriculum (among other things, MATS students attended a course on markup languages, which ran parallel to the AI course and they will attend a separate course on Python programming in the second semester). Since attrition was considerably higher among MAFKÜ students compared to MATS students (58.33\% vs. 33.33\%), the post-test group likely exhibits a higher baseline of IT competence and technical disposition than the initial pre-test group. This is supported by the participants' self-assessment related to IT competence (item 0.6) and interest in the technical foundations of language-oriented AI (item 0.5). The share of participants reporting high or very high IT competence almost doubled from 20.9\% in the pre-test to 40\% in the post-test. Similarly, the share of participants reporting a high or very high interest in the technical foundations of language-oriented AI increased from 70.8\% in the pre-test to 80\% in the post-test.\footnote{There may be a combination of a selection effect (higher retention rate of students with a higher initial IT affinity; [very] high IT competence pre-test MAFKÜ group: 8.33\% vs pre-test MATS group: 22.22\% vs. pre-test professionals: 66.67\%) and a treatment effect (increase in IT competence through contents covered in the technical curriculum) at work here. This will be discussed in more detail in Section 3.1.} 

\subsection{Overall didactic effectiveness of the curriculum}
The participants' language-oriented AI knowledge was measured using three questionnaire items (items 1.1 to 1.3) with an 11-point Likert scale ranging from 0 = \textit{Hardly/Not at all} to 10 = \textit{Perfectly} without intermediate scale labelling.\footnote{\textit{I can explain the basics of how modern language-oriented AI technologies work} (item 1.1), \textit{I can explain how modern language-oriented AI technologies are trained and fine-tuned} (item 1.2), \textit{I know if language-oriented AI supports the translation technologies I use and, if so, how} (item 1.3).} The questions were asked in both the pre- and the post-test in order to measure participants' self-reported knowledge gain. Results are presented in table 1.

\begin{table*}[t]
\centering
\label{tab:knowledge_gain}
\begin{tabular}{lcccc}
\hline
Knowledge area & Pre-test ($M \pm SD$) & Post-test ($M \pm SD$) & $p$-value & Cohen's $d$ \\ 
\hline
1.1 Basic operating principle & $3.67 \pm 2.24$ & $6.73 \pm 1.33$ & $< 0.001$ & 1.58 \\
1.2 Training/Fine-tuning & $3.04 \pm 2.18$ & $5.87 \pm 1.77$ & $< 0.001$ & 1.39 \\
1.3 AI-supported trans tech & $4.46 \pm 2.77$ & $7.67 \pm 2.09$ & $< 0.001$ & 1.27 \\
\hline
\textbf{Composite index} & $\mathbf{3.72 \pm 2.13}$ & $\mathbf{6.76 \pm 1.43}$ & $\mathbf{< 0.001}$ & \textbf{1.60} \\ 
\hline
\multicolumn{5}{l}{\small $N_{Pre} = 24, N_{Post} = 15$, 11-point Likert scale from 0 (\textit{Hardly/Not at all}) to 10 (\textit{Perfectly})}
\end{tabular}
\caption{Self-reported knowledge gain of participants with regard to the technical foundations of language-oriented AI}
\end{table*}

We focus here on the composite index, which represents the mean of the responses to items 1.1 to 1.3 for each participant (serving as an aggregate of participants' self-assessed knowledge of the technical foundations of language-oriented AI). The increase in participants' self-assessed technical knowledge from $M = 3.72$ ($SD = 2.13$) in the pre-test to $M = 6.76$ ($SD = 1.43$) in the post-test is highly statistically significant ($p \textless 0.001$)\footnote{Welch’s t-test for independent samples was used to account for the lack of consistent participant IDs across pre- and post-test and to address unequal group sizes and variances (\textit{scipy.stats.ttest\_ind}, \textit{equal\_var=False}).}, with a very large effect size of $d = 1.60$. This suggests a high overall didactic effectiveness of the technical curriculum, taking into account that generalizability may be limited to some extent due to the relatively small sample size.\\
The post-test questionnaire also included a retrospective question where participants were asked to reassess their technical knowledge of language-oriented AI at the beginning of the course.\footnote{\textit{Looking back at the beginning of the course, how would you now assess your knowledge of the technical foundations of language-oriented AI at that time?} (item 1.4, 11-point likert scale ranging from 0 = \textit{Hardly any/No knowledge} to 10 = \textit{Complete/Perfect knowledge})} This was done to account for a potential Dunning-Kruger effect at the onset of the study. The results of this retrospective self-assessment ($M = 2.93$, $SD = 2.19$) were notably lower than the initial self-assessment ($M = 3.72$, $SD = 2.13$), suggesting a response shift. As the curriculum introduced participants in detail to  the technical foundations of language-oriented AI, they likely became aware of previous knowledge gaps that may have skewed their initial self-assessment. This indicates that the knowledge gain associated with the technical curriculum was even more pronounced than the initial pre-post comparison suggests. The response shift may also indicate an increase in participants' metacognitive awareness in terms of a more realistic picture of their own level of competence with regard to language-oriented AI.\footnote{Metacognition has also been discussed as a dimension of AI literacy acquisition in that it ``[enables] individuals to reflect on their learning processes and understand how AI can enhance their cognitive abilities.`` \cite[3]{Tadimalla_Maher_2024}}\\
As mentioned above, the interpretation of results must take into account a potential selection effect (students with higher initial IT affinity completing the course and students with lower initial IT affinity leaving the course). However, both the observed response shift and the very large effect size of $d = 1.60$ (or $d = 2.07$ for the retrospective self-assessment of $M = 2.93$, $SD = 2.19$) provide evidence for a genuine treatment effect. These factors indicate that the self-reported knowledge gain indeed resulted from participants' engagement with the technical topics covered in the curriculum, rather than being merely an effect of participant self-selection.    
\subsection{Didactic potential of Jupyter notebooks}
The post-test questionnaire also asked participants to assess the didactic potential of Jupyter notebooks in the context of the technical curriculum. When participants were asked whether these notebooks were a suitable didactic instrument for covering the technical course contents (item 2.1), 12 (80\%) strongly agreed, 1 (6.7\%) agreed and 2 (13.3\%) neither agreed nor disagreed. This largely confirms the results reported in Krüger \shortcite[519--520]{Krueger_2022} concerning the didactic potential of Jupyter notebooks in translation technology teaching. When asked specifically whether the combination of multimodal documentation (text + figures + linked videos) and executable code cells helped participants to understand the technical contents of the curriculum (item 2.2), 11 (73.3\%) strongly agreed, 3 (20\%) agreed and 1 (6.7\%) neither agreed nor disagreed, which can be taken as further evidence in favour of the notebooks' didactic suitability.\\
Participants in the post-test study could also provide open comments on the didactic potential of the Jupyter notebooks for covering the technical course contents (five participants provided such comments). Overall, feedback was positive. E.g., according to a MATS student, the notebooks \textit{illustrated the content very clearly and generally helped me to review the lecture material at my leisure (the notebooks were also very detailed, which was also helpful)}.\footnote{Participants provided all open comments in German. The comments were machine-translated into English using DeepL and post-edited by the author.} A MAFKÜ student found \textit{the structure and detail of the descriptions/explanations very helpful. Better than slides with bullet points, because it gives you the opportunity to read up on something that has already been explained in detail by the lecturer in the seminar}. The same student reported having difficulties saving the notebooks locally (which could have been achieved through adequate lecturer support). Another MAFKÜ student made an interesting point regarding further didactic scaffolding in addition to the notebooks: \textit{In the context of our course and your explanations of the content of the lines of code, I found these very helpful. However, I'm not so sure whether I—or other users who have no prior knowledge of coding—would have been able to learn well with these alone, especially since the relevant information was often hidden in long lines of code}. This indicates that, in the zone of proximal development, expert guidance cannot be fully embedded into the notebooks themselves and should also be provided by the lecturer (in this specific case by highlighting relevant portions of longer and potentially complex code cells). Future iterations of the AI course will also introduce students to using LLMs as coding assistants for producing, highlighting, explaining, changing or debugging computer code, as discussed in a translation context by Krüger et al. \shortcite{Krueger_et_al_forthcoming}. Such LLM integration may offer further didactic potential, e.g., in terms of using these models as general learning assistants facilitating deliberate practice \cite{Angelone_2026} in addition to lecturer support or in self-learning scenarios where no such lecturer support is available.\footnote{LLM's potential as learning assistants may be further reinforced by the possibility to prompt these models to tailor the ``level of explanatory ambition`` \cite[82]{Engberg_2025} to individual knowledge requirements (e.g., explaining the concept of a backward pass through a neural network to MSc students in information technology vs. MA students in translation/specialised communication).}    

\subsection{Further feedback on the technical curriculum}
When asked whether the progression of the course contents from 1) vector embeddings via 2) the technical foundations of neural networks and 3) subword tokenization to 4) transformer neural networks  was suitable from a didactic point of view (post-test questionnaire, item 3.1), 12 participants (80\%) found the order of topics highly suitable and 3 (20\%) found it rather suitable. In an open comment, one of the professional participants noted that they would have liked a final overview of how the different AI concepts covered in the course relate to each other. This will be implemented in a future version of the curriculum.\\
Finally, participants were asked to provide concluding feedback (in the form of open comments) on the AI course and the technical curriculum, e.g., in terms of the usefulness of the course in the context of their MA programme or with a view to participants' (future) careers, in terms of the appropriateness of the technical level at which the content was covered or in terms of any content that was missing or was covered too briefly or superficially (item 4.1). Seven participants provided concluding feedback.\\
The technical level was considered demanding (particularly in the context of transformer language models), but participants largely agreed that it could be mastered using the didactic resources provided. A MAFKÜ participant noted: \textit{Although I consider the course to be challenging, I find that the content is explained very clearly and from a perspective that is easy for me to understand (as a translator)}. A professional translator participant commented that, at times, there was a risk of getting lost in too much technical detail. Another MAFKÜ participant responded that, although the lecturer always encouraged students to ask questions if they did not understand something, it was sometimes difficult to put such questions into words in the first place. These issues will have to be addressed in future iterations of the course, also taking into account the didactic potential of LLMs as coding or general learning assistants as discussed above.\\
Overall, participants acknowledged the relevance of the course and the curriculum for their (later) professional career. A MAFKÜ student noted that even if a technical understanding of language-oriented AI was not directly relevant in students' future careers, \textit{I believe that this background knowledge gives participants a certain degree of confidence}. A professional translator noted that security restrictions currently prevented them from using language-oriented AI technologies extensively at their workplace but that \textit{I can at least discuss with our IT department what is important for language service providers, make suggestions and discuss things on an equal footing}. These comments support the hypothesis stated in Section 1 that technical AI literacy as acquired through the technical curriculum includes an empowerment dimension and may strengthen L\&T stakeholders' (algorithmic) agency (e.g., by discussing language-oriented AI `on an equal footing` with IT experts and/or consulting on workflow design) and may contribute to their overall digital resilience (in terms of confidence) in highly automated environments.  

\section{Conclusion and outlook}
This paper presented a technical curriculum on language-oriented AI in translation and specialised communication, which stands in the tradition of other L\&T stakeholder-specific initiatives such as MultiTraiNMT, DataLit\textsuperscript{MT}, adaptMLLM and LT-LiDER. The overall didactic suitability of the curriculum was tentatively confirmed in an AI-focused MA course at TH Köln. Participants acknowledged the relevance of the course and the technical curriculum for their (later) professional careers, with individual participant feedback supporting the hypothesis that technical AI literacy may strengthen L\&T stakeholders' (algorithmic) agency and digital resilience in professional high-automation contexts. Participants also confirmed the didactic potential of Jupyter notebooks for covering the technical course contents. However, participant feedback also indicated that the technical content could be overwhelming at times or was only manageable through additional lecturer support. This indicates that, in the zone of proximal development, where learners are challenged by the novelty and/or complexity of content, the required expert guidance cannot be fully embedded into the Jupyter notebooks and should also be provided through higher-level didactic scaffolding, ideally in terms of lecturer support. Participant feedback will be implemented into the curriculum on an ongoing basis in order to tailor it further to the specific knowledge requirements of the target audience. This is intended to strengthen the curriculum's \textit{human-centered explainable AI} (HCXAI) dimension, which aims to place non-expert users at the center of AI explainability \cite[98]{Ridley_2025}. Also, future work will investigate strategies for introducing LLMs as coding or general learning assistants into the curriculum, which could facilitate deliberate practice in self-learning scenarios.    

\paragraph{Acknowledgements:} The author would like to thank his colleagues Janiça Hackenbuchner and Erik Angelone for their valuable comments and suggestions on the manuscript version of this paper.

\bibliography{eamt26}
\bibliographystyle{eamt26}
\end{document}